\pdfoutput=1

\documentclass[11pt]{article}

\usepackage[final]{acl}
\usepackage{xcolor}
\usepackage{times}
\usepackage{latexsym}
\usepackage[labelformat=simple]{subcaption}

\usepackage{algorithm}  
\usepackage{algorithmicx}  
\usepackage[noend]{algpseudocode}
\usepackage[T1]{fontenc}
\usepackage{multirow}

\usepackage[utf8]{inputenc}

\usepackage{colortbl}

\usepackage{makecell}

\usepackage{microtype}
\usepackage{amsmath}
\usepackage{array,amsfonts}
\usepackage{inconsolata}
\usepackage{graphicx}
\graphicspath{ {images/} }
\usepackage{enumitem}

\setlist[itemize]{itemsep=1.5pt}

\DeclareGraphicsExtensions{.pdf,.png,.jpg,.gif,}
\title{An Entropy-based Text Watermarking Detection Method}

\author{Yijian LU\textsuperscript{1}, Aiwei Liu\textsuperscript{2} , Dianzhi Yu\textsuperscript{1} , Jingjing Li\textsuperscript{1} , Irwin King\textsuperscript{1}\thanks{Corresponding Author.} \\
        \textsuperscript{1}The Chinese University of Hong Kong \\\textsuperscript{2}Tsinghua University \\
\small{\texttt{luyijian@link.cuhk.edu.hk}}, \small{\texttt{liuaw20@mails.tsinghua.edu.cn}} \\
\small{\texttt{dianzhi.yu@link.cuhk.edu.hk}}, \small{\texttt{lijj@link.cuhk.edu.hk}}, 
\small{\texttt{king@cse.cuhk.edu.hk}}}

\begin{document}
\maketitle

\begin{abstract}
Text watermarking algorithms for large language models (LLMs) can effectively identify machine-generated texts by embedding and detecting hidden features in the text. Although the current text watermarking algorithms perform well in most high-entropy scenarios, its performance in low-entropy scenarios still needs to be improved. In this work, we opine that the influence of token entropy should be fully considered in the watermark detection process, $i.e.$, the weight of each token during watermark detection should be customized according to its entropy, rather than setting the weights of all tokens to the same value as in previous methods. Specifically, we propose \textbf{E}ntropy-based Text \textbf{W}atermarking \textbf{D}etection (\textbf{EWD}) that gives higher-entropy tokens higher influence weights during watermark detection, so as to better reflect the degree of watermarking. Furthermore, the proposed detection process is training-free and fully automated. From the experiments, we demonstrate that our EWD can achieve better detection performance in low-entropy scenarios, and our method is also general and can be applied to texts with different entropy distributions. Our code and data is available\footnote{\url{https://github.com/luyijian3/EWD}}. Additionally, our algorithm could be accessed through MarkLLM \cite{pan2024markllm}\footnote{\url{https://github.com/THU-BPM/MarkLLM}}.

\end{abstract}

\section{Introduction}

\begin{figure}[t]
\includegraphics[width=0.49\textwidth,height=0.31\textwidth]{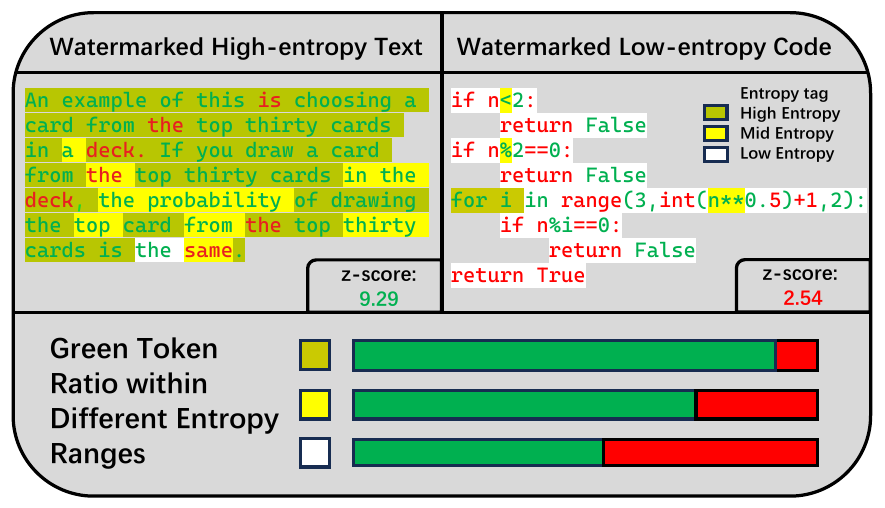}
\caption{\label{intro-fig}This figure shows that compared with watermarked texts with mostly high-entropy tokens, watermarked codes with mostly low-entropy tokens see significantly less green tokens, resulting in a small detection z-score. Furthermore, on the bottom of the figure, we demonstrate that the green token ratio in tokens decreases as their entropy decreases.}


\end{figure}

The rapid advancements in large language models (LLMs) have enabled them to generate high-quality outputs indistinguishable from humans, and also achieve better performance on various real-world generation tasks. However, this gives rise to the potential risk of misuse. Specifically, LLM-generated assignments, including essays and codes, pose a threat to academic integrity. Misusing LLMs to generate fake news can also lead to negative societal consequences \cite{megias2021dissimilar}. Therefore, algorithms capable of effectively identifying machine-generated contents have become essential.

Text watermarking algorithms can effectively alleviate the problem of LLM misuse by embedding and detecting hidden features in LLM-generated contents. 
For example, the watermarking algorithm proposed by  \citet{DBLP:conf/icml/KirchenbauerGWK23} (KGW)  divides the LLM's model vocabulary into two lists, green and red, and increases the logits value of tokens in the green list by a positive constant (watermark strength). 
The modified logits will bias the LLM to sample most tokens from the green list.
As a result, the watermarked text containing a dominant number of green tokens will be recognized by the watermark detector while human texts will not. However, the successful watermark generation is largely tied to the token entropy during text generation. Under the watermark algorithm, the probability of sampling a token from the green list is proportional to its entropy, meaning that low-entropy tokens would produce far less green tokens than high-entropy tokens, as illustrated by Figure~\ref{intro-fig}. In that case, green tokens will occupy only half of a watermarked low-entropy text, resulting in a smaller detection score and being mistakenly labeled as human-written. Furthermore, to ensure the text quality of output contents, the watermark strength in the generator cannot be excessively enlarged to forcefully modify the logits and sample more tokens from the green list \cite{tu2023waterbench}. 



Consequently, a crucial problem is how to improve the watermark detector so that low-entropy watermarked texts with a limited number of green tokens can be successfully identified. \citet{lee2023wrote} propose a selective watermark detector named SWEET to better reflect the watermark level in low-entropy texts. Specifically, they only detect tokens that exceed a certain entropy threshold, thus excluding tokens that are difficult to be affected by the watermark generator. However, this method itself has three weaknesses. First, their method requires a manually-set entropy threshold by analyzing samples of code written by humans. Second, their method also does not consider the distribution of token entropy, that is, their method treats all tokens that exceed the threshold as the same, while in fact the entropy of these tokens may vastly vary. Third, the performance gain using the SWEET detector is limited and there is still significant room for further advancement.

To better address the low-entropy watermarking detection problem than previous methods,
we introduce an \textbf{E}ntropy-based Text \textbf{W}atermark \textbf{D}etection (\textbf{EWD}) algorithm to 
fully consider the role of entropy in the detection of a given text. Specifically, we propose that the influence of a token during detection should be proportional to its entropy. 
Furthermore, in order to fully reflect this positive relation between token entropy and detection influence, we utilize a monotonically-increasing and continuous function to generate influence weights from token entropy.
By creating a gap between the weights of high entropy tokens and low entropy tokens, we ensure that if the watermark generation algorithm can hardly affect a low-entropy token, the state of that token will make minimal impact to the detection result. Overall, our method has three major benefits. First, the method saves the trouble of altering the watermark generation process. Second, compared to SWEET, our method is more efficient as the acquisition of token weights is an independent and training-free process, avoiding the trouble of a human dataset or any manual operations. Third, the proposed detection scheme can be applied to machine-generated texts of all entropy levels, not solely low-entropy texts.



From the experiments, we demonstrate that the detection accuracy of EWD in low-entropy scenarios surpasses existing baselines. Furthermore, EWD exhibits versatility in detecting texts of high entropy levels, as well as robustness against the back-translation  watermark-removal attack.

In summary, the contributions of our work are summarized as follows:

\begin{itemize}
  \item We propose an entropy-based text watermarking detection method called \textbf{EWD}. 
  \item We provide a theoretical analysis of the detection accuracy in the low-entropy setting.
  \item We show that EWD not only improves the detection accuracy under the low-entropy scenario, but also provides similar performance as other methods in high-entropy texts.
\end{itemize}

\section{Related Work}
In this section, we aim to review existing works on text watermarking algorithms. Overall, text watermarking can be divided into two different categories: watermarking for existing texts and watermarking for LLMs \citet{liu2023survey}.

\noindent\textbf{Text Watermarking for Existing Texts.}
Watermark message can be embedded into existing text by performing a series of modifications, such as lexical substitutions \cite{munyer2023deeptextmark,topkara2006hiding,yang2022tracing,yoo2023robust,yang2023watermarking}. Those methods typically replace words with their alternatives to place watermark. For example, \citet{topkara2006hiding} used WordNet\cite{fellbaum1998wordnet} as his dictionary while \citet{munyer2023deeptextmark} utilized a pretrained Word2Vec model to select the alternative words. To ensure that the repalced words would cause minimal changes to the text semantics, 
\citet{yang2022tracing} proposed a BERT-based infill model to generate replacement with regard to the sentence meaning. To improve the robustness of the algorithm against possible corruptions, \citet{yoo2023robust} fine-tuned the infill model to generate consistent word distribution given several corrupted texts of the same origin. However, the level of modification to an existing text is often limited to protect text quality in the watermarked output. Hence, watermarking for existing text is less favored due to its limited capacity and effectiveness.

\noindent\textbf{Text Watermarking for LLMs.} Watermark embedding can also be conducted during the LLM generation phase, either by modifying the token sampling or model logits. \citet{christ2023undetectable} introduced a technique that incorporates watermark message by predefining a random number sequence for sampling the tokens. 
To enhance resistance to text modifications, \citet{kuditipudi2023robust} utilized the Levenshtein distance in matching the text and numbers. For the latter, the most representative one that modifies the model logits is proposed by \citet{DBLP:conf/icml/KirchenbauerGWK23} based on previous tokens during text generation. 
There are a number of works proposed on top of this algorithm with enhanced performance in payload \cite{yoo2023advancing,wang2023towards}, robustness \cite{zhao2023provable,liu2024a,kirchenbauer2023reliability,ren2023robust,he2024can} and unforgeability \cite{liu2024an,cryptoeprint:2023/1661}. For example, \citet{yoo2023advancing} extends the method to a multi-bit setting.
\citet{wang2023towards} proposed to utilize a proxy LLM for multi-bit message embedding. To increase robustness against editing attacks, \citet{zhao2023provable} proposed to set a universal green list during the generation process. \citet{liu2024a} proposed a different approach called SIR which utilizes the semantic meaning of previous generated text to determine the green lists. Inspired by \citet{liu2024a}, \citet{he2024can} firstly proposed the Crosslingual Watermark Removal Attack (CWRA), then offered X-SIR as a defense method against CWRA. Also, to enhance the unforgeability of text watermark, 
\citet{liu2024an} introduced a neural network for text generation and detection. In this work, we focus only on the detection of low-entropy texts. \citet{lee2023wrote} designed a selective watermark generation and detection approach for low-entropy code generation. However, it requires a human code dataset to help manually determine the entropy threshold. To sum up, there lacks an effective method to improve the low-entropy detection problem.

\section{Preliminaries}
\subsection{Text Generation Process of LLMs}
Here we introduce the necessary concepts used in this paper. A LLM, $M$ takes a prompt as input and outputs subsequent tokens as the corresponding response. Specifically, we denote the initial prompt as $x^{prompt}$. Then, at the $l$-th step, the input to the LLM is the $x^{prompt}$ and the already-generated tokens $T_{:(l-1)}$. The LLM would generate a distribution logits $P_{LLM}(x^{prompt},T_{:(l-1)})$, consisting of  $P_{LLM}(v|x^{prompt},T_{:(l-1)})$ over each token $v$ of its model vocabulary  based on the joint input. 

\subsection{Text Watermarking}
The general watermarking scheme used in this paper is called KGW, which is to modify the logits generation process of an original LLM $M$ to acquire the watermarked LLM, denoted as $\tilde{M}$. The modification on the original logits by adding a small value (the watermark strength $\delta$) to the logits of green-list tokens will result in LLM's preference on those tokens in the output text. Thus such modification could likely produce a green token. 

The detection algorithm is to calculate the watermark level using the following equation,
\begin{equation}
\label{z-score}
z = (|s|_G - \gamma |T|) / \sqrt{\gamma(1-\gamma)|T|},
\end{equation}
where $|s|_G$ is the number of green tokens in text $T$ and $\gamma$ is the green-list ratio in the model vocabulary. The algorithm would output $Detect(T) = 1$ if the value is greater than a given threshold, and $Detect(T) = 0$ otherwise.

\subsection{Token Entropy and Low-entropy Scenario}
The entropy of a token is measured by the degree of spread-out of the logits generated by the LLM during the sampling process of that token. In this paper, we follow the spike entropy from KGW, 
\begin{equation}
    SE(k) = \sum_{v}{\frac{p_v}{1+\tau p_v}},
\end{equation}
where $p_v$ is the logits value of each token $v$ in the vocabulary when sampling the token $k$, and $\tau$ is a scalar. The entropy value reaches the minimal value if the logits concentrates on a single token and the maximal value given a uniform distribution.

The watermark generation process is more effective on a high-entropy token where the logits distribution is nearly uniform, so that the modification could produce a green-list token easily. On the other hand, a low-entropy token is almost deterministic and hard to be affected. Therefore, the green-list ratio for a low-entropy token is much lower, resulting in fewer green tokens in the output. Therefore, low-entropy watermarked texts pose threats on the detection accuracy as they are easily identified as human-written.












\section{Proposed Method}

In this section, we will first explain the motivation by analysing the watermark generation effect on low-entropy tokens, then provide a detailed explanation of the proposed EWD. 


\subsection{Motivation}

Entropy plays an important role in the watermark generation process, yet its role in the detection process has been underestimated. In the KGW generator, a positive constant is added to the logits value of all green-list tokens.
The distribution level of the logits, represented by the entropy, matters significantly during this process. When the logits resembles a uniform distribution,
this logits modification would very likely produce a green-list token. 
However, when the logits concentrates on one single token,
this watermark embedding process would hardly affect the token sampling and produce a green token with a much lower probability close to the green list ratio $\gamma$.
Since entropy determines the watermark generation process, then it should also affect the watermark detection. However, in the KGW detector, all tokens are of the same influence, no matter how their entropy varies.


\subsection{Entropy-based Text Watermarking Detection}
Our proposed EWD aims to fully consider the role of entropy in the watermark detection process. Specifically, we firstly propose a positive influence-entropy relation, then utilize entropy for detection weight customization.

The influence of a token during detection should be proportional to its entropy. From previous sections, we explain why a high-entropy token is easier to be affected by the watermark generator, as well as the situation for a low-entropy token. Since a low-entropy token is hard to be affected, then whether it is sampled from the green list or not should have a limited influence on the overall watermark level. Similarly, when an easy-to-affect (high-entropy) token is not green, it should contribute more to showing that the text $T$ is not watermarked. Therefore, we introduce a positive influence-entropy relation to determine how much a token $t$ can influence the detection result:
\begin{equation}
\label{relation}
    I(Detect(T) = \textbf{1}(t\in G)|t) \propto SE(t),
\end{equation}
where $T$ is the given text and $G$ is the green list. 



The detection weight of a token should be customized based on its entropy to fully reflect the positive influence-entropy relation. The influence index in Eq.~(\ref{relation}) is represented by the detection weight of each token. However, the detection weight assigning of each token in SWEET fails to reflect the positive influence-entropy relation. This is because that inside the two groups of above-threshold tokens and below-threshold tokens, the weights of tokens with different entropy are identical.
Therefore, to fully reflect the influence-entropy relation, we utilize a monotonically-increasing and continuous function $f$ to customize weights from token entropy. The weight $W(t)$ of a token $t$ can be illustrated as follows:\begin{equation}
\label{assignment}
    W(t) = f(SE(t) - C_0),
\end{equation}
where $C_0$ is the minimal value of the spike entropy to normalize the entropy input before computing the weight. Using the entropy-based weight customization, even a sight increase in the token entropy will result in a increased detection weight. Table~\ref{table_weight} summarizes the weight features of different methods and their reflection of the proposed influence-entropy relation. 

\begin{table}[t]
\small
\caption{Features of detection weights by different methods and how they can reflect the proposed positive influence-entropy relation.}\label{table_weight}
\centering
\begin{tabular}{ccc}
\hline
Methods & Detection Weight & \begin{tabular}[c]{@{}c@{}}Reflection of\\ Eq.~(\ref{relation})\end{tabular} \\ \hline
KGW     & Constant       & None                                                               \\
SWEET   & Binary         & Partial                                                             \\ \rowcolor{gray!20}
EWD     & Customized     & Full                                                               \\ \hline
\end{tabular}
\end{table}

\begin{algorithm}[t]
\small
\caption{Entropy-based Watermark Detection}\label{alg_detect}
\begin{algorithmic}
\State \textbf{Input:}{ a language model $M$, input text $T$, detection key 

$k$, window size $m$, detection threshold $\tau$, green list 

ratio $\gamma$}

\State {Compute token entropy based on model logits, 

$SE=SpikeEntropy[M(T)]=[SE_0,SE_1....]$}
\State {Compute token weights based on token entropy, 

$W=ComputeWeight[SE]=[W_0,W_1....]$}
\For {$l=m,m+1....$}
\State{Use the detection key $k$ and previous $m$ tokens 

$T_{l-m:l-1}$ to find the green list $G$}
\State{If current token $T_l$ is inside the green list $G$, then 

add its weight $W_l$ to $|s|_G$}
\EndFor
\State {Compute the detection score $z'$ by Eq.~(\ref{z-score-ours})}
\If{$z'>\tau$} \textbf{return} 1, $i.e.,$ "Input text is watermarked"
\Else{} \textbf{return} 0, $i.e.,$ "Input text is not watermarked"
\EndIf
\end{algorithmic}
\end{algorithm}

Pseudocode of our proposed detection method is provided in Algorithm~\ref{alg_detect}. We first acquire the text entropy by computing the model logits of each token. The next step is to determine the influence weight of each token, the output of a designed function \textit{ComputeWeight} with their entropy values as inputs. 
After obtaining the entropy-based weights for each token, we then apply the standard detection procedure by \citet{DBLP:conf/icml/KirchenbauerGWK23} to determine the group of green tokens, $i.e.$, computing the green lists using the detection key and previously generated tokens. Finally, we sum up the weights of the green tokens $|s|_G$ and calculate the z-score by
\begin{equation}\centering
\label{z-score-ours}
z' = (|s|_G - \gamma \sum_{i = m}^{|T|-1}{W}_i) / \sqrt{\gamma(1-\gamma)\sum_{i = m}^{|T|-1}{W_i}^2},
\end{equation}
given $|s|_G$ and weight $W_t$ of each detected token $t$ in text $T$. The detector would give a positive result if the z-score is greater than the given threshold. It is noticably that the proposed algorithm is a fully automated process that eliminates the need for manual operations from the SWEET method.

\section{Theoretical Analysis}
In this section, we theoretically analyze the Type-I and Type-II error of our proposed detection method. Specifically, we compare EWD with two previous detection methods, KGW 
 and SWEET.

To do so, we adopt the same steps as in KGW \cite{DBLP:conf/icml/KirchenbauerGWK23}. Each token in both watermarked and non-watermarked texts can either be green or red, so we can model it as a binomial distribution. While \citet{zhao2023provable} set the window size in the watermark generator as zero (one universal green list for all tokens) and exhibits major token dependence, $i.e.$, the same tokens in the text are of the same color, in our case, where a non-zero window size is adopted, the token dependence is significantly weakened. For example, the same token with different prefix would exhibit different sampling result. 

Therefore, to simplify and facilitate the process of theoretical analysis, we make the reasonable assumption that the sampling of each token is independent. Hence, the sum of green tokens in the text can be approximated with a normal distribution $\mathcal{N}(\mu, \sigma^2)$ with different sets of mean $\mu$ and variance $\sigma^2$ for different detection schemes. Finally, we calculate and compare the probability of the sum surpassing or falling behind the detection threshold $|\tilde{s}|_G$, which is the $|s|_G$ value in each detection function with a fixed $z$.

\subsection{Type-I Error}
Type-I error measures the likelihood a human-written text is incorrectly identified as being generated by a watermark algorithm. 

In a human-written text $T$, each token is independent of the watermark algorithm. Therefore, its probability to be included in the green list is $\gamma$. Therefore, KGW's sum of green tokens is approximated by $\mathcal{N}(\gamma |T|, \gamma(1-\gamma) |T|)$. For SWEET that only considers high-entropy tokens, such distribution is modelled as $\mathcal{N}(\gamma \tilde{|T|}, \gamma(1-\gamma) \tilde{|T|})$, where $\tilde{T}$ is the text excluding low-entropy tokens. Our method EWD assign weights to tokens based on entropy, leading to a weighted binomial distribution. We could also approximate it to a normal distribution 
with mean $\mu=\gamma \sum_{i=m}^{|T|-1}{W}_i $ and variance $ \sigma^2=\gamma(1-\gamma)\sum_{i=m}^{|T|-1}{W_i}^2$.


The theoretical Type-I errors for all detection methods are the same. When the detection threshold is set to 2, the probabilities for $|s|_G$ under each method to surpass its detection threshold $|\tilde{s}|_G$ is identically 2.28\%.




\subsection{Type-II Error}
Type-II error measures the likelihood a machine-generated text is incorrectly identified as human-written. 
For analysis, we assume that a sequence generated by the watermark algorithm has a length of $|T| = 200$ tokens and the watermark algorithm has hyper-parameters $\gamma = 0.5$ and $\delta = 2$.

Originally, KGW samples over 500 watermark-generated news scripts and reports an average spike entropy of 0.807. In this section, we focus specifically on cases with a much lower average spike entropy.
Based on our data, 
we approximate the distribution of spike entropy in low-entropy watermarked texts with a power law distribution $f(x) = a x^{a-1}$, where the parameters are $a = 0.106$, $loc = 0.566$, $scale = 0.426$. 
Under this distribution, the average spike entropy is 0.608.
The following probability from KGW shows a lower bound of the probability of a token $k$ being sampled from the green list:
\begin{equation}
\label{kgw_prob}
    \mathbb{P}[k \in G] \geq \frac{\gamma \alpha}{ 1 + (\alpha - 1)\gamma} SE(k),
\end{equation}
where $\alpha = exp(\delta)$. We would utilize this probability later in the calculation of $\mu$ and $\sigma^2$.



\textbf{KGW.} Based on Eq.~(\ref{kgw_prob}), the statistical $\mu$ and upper bound of $\sigma^2$ can be obtained as $\sum_{k}\mathbb{P}[k\in G]$ and $\sum_{k}\mathbb{P}[k\in G]\cdot(1-\mathbb{P}[k\in G])$. When the average entropy $SE^*$ given by the simulated power law distribution is 0.608, we approximate $|s|_G$ by $\mathcal{N}(107.10, 49.70)$. If we set a threshold of $z$ = 2 (corresponding to $|\tilde{s}|_G$ = 114.14) for detection, the false negative rate exceeds 84.1\%. 
This means that KGW is not good at detecting the watermarked texts in cases of low entropy. 

\textbf{SWEET.} Following SWEET, we set the entropy threshold at 0.695 and exclude all tokens below this threshold during detection. The average spike entropy $SE^*$ and remaining text length $\tilde{|T|}$ then change to 0.82 and 24, respectively. Using the equations from KGW, we can get the approximation as $\mathcal{N}(17.48, 4.84)$. If we set a threshold of $z$ = 2 (corresponding to $|\tilde{s}|_G$ = 17.02) for detection, the false negative rate exceeds 41.7\%.


\textbf{EWD.} EWD assigns higher influence weights to tokens with high spike entropy.  
Note that 
in experiments, we use several weight functions, and 
here, for analysis purposes, we use a linear weight function for each token $k$: $W(k) = SE(k)-C_0$, where $C_0$ is a constant. Based on Eq.~(\ref{kgw_prob}), the statistical $\mu$ and $\sigma^2$ can be obtained as $\sum_{k}\mathbb{P}[k\in G]\cdot W(k)$ and $\sum_{k}{W(k)}^2\mathbb{P}[k\in G]\cdot (1-\mathbb{P}[k\in G])$. Therefore, we can approximate the $|s|_G$ detected by EWD with $\mathcal{N}(5.79, 0.31)$. If we set a threshold of $z'$ = 2 (corresponding to $|\tilde{s}|_G$ = 5.55) for detection, the false negative rate exceeds 33.4\% (details in Appendix~\ref{appendix:1}). 

Table~\ref{table_error} summarizes our calculated theoretical Type-I and Type-II error using different methods in low-entropy detection scenarios. Overall, our method achieves the lowest Type-II error among all methods while maintaining the same Type-I error.

\begin{table}[t]
\small
\centering
\caption{This table summarizes our calculated theoretical Type-I and Type-II error using different methods in low-entropy detection scenarios. The data in the table is presented in percentages (\%). Lower values indicates better detection accuracy.}\label{table_error}
\begin{tabular}{ccc}
\hline
Methods       & Type-I Error   & Type-II Error \\ \hline
KGW        & 2.28    & 84.1     \\
SWEET             & 2.28    & 41.7    \\ \rowcolor{gray!20}
EWD          & \textbf{2.28}    & \textbf{33.4}    \\ \hline
\end{tabular}
\end{table}

\section{Experiments}
\label{sec:Experiments}

\begin{table*}[t]
\centering
\caption{Main results on different datasets using KGW, SWEET and our EWD for detection.}\label{table_main}
\begin{tabular}{cccccclccccc}
\hline
\multicolumn{12}{c}{Code Detection (Low-entropy Scenario)}                                                                                                            \\ \hline
        & \multicolumn{5}{c}{HumanEval}                                  &  & \multicolumn{5}{c}{MBPP}                                        \\ \cline{2-12} 
Methods & \multicolumn{2}{c}{1\%FPR} & \multicolumn{2}{c}{5\%FPR} & Best  &  & \multicolumn{2}{c}{1\%FPR} & \multicolumn{2}{c}{5\%FPR} & Best  \\ \cline{2-12} 
        & TPR          & F1          & TPR          & F1          & F1    &  & TPR          & F1          & TPR          & F1          & F1    \\ \hline
KGW     & 0.331        & 0.494       & 0.414        & 0.567       & 0.772 &  & 0.115        & 0.205       & 0.348        & 0.498       & 0.744\\
SWEET   & 0.455        & 0.622       & 0.667        & 0.778       & 0.842 &  & 0.409        & 0.577       & 0.650        & 0.765       & 0.865 \\
 \rowcolor{gray!20} 
EWD    & \textbf{0.466}        & \textbf{0.633}       & \textbf{0.692}       & \textbf{0.797}       & \textbf{0.859} &  & \textbf{0.567}        & \textbf{0.720}       & \textbf{0.744}        & \textbf{0.830}       & \textbf{0.878} \\ \hline
\multicolumn{12}{c}{Rotowire (High-entropy Scenario)}                                                                                                   \\ \hline
        & \multicolumn{5}{c}{Sampling}                                    &  & \multicolumn{5}{c}{Beam-search}                                 \\ \cline{2-12} 
Methods & \multicolumn{2}{c}{1\%FPR} & \multicolumn{2}{c}{5\%FPR} & Best  &  & \multicolumn{2}{c}{1\%FPR} & \multicolumn{2}{c}{5\%FPR} & Best  \\ \cline{2-12} 
        & TPR          & F1          & TPR          & F1          & F1    &  & TPR          & F1          & TPR          & F1          & F1    \\ \hline
KGW     & 1.000        & 0.995       & 1.000        & 0.976       & 1.000 &  & 1.000        & 0.995       & 1.000        & 0.976       & 1.000 \\
SWEET   & 1.000        & 0.995       & 1.000        & 0.976       & 1.000 &  & 1.000        & 0.995       & 1.000        & 0.976       & 1.000 \\
 \rowcolor{gray!20} 
EWD    & \textbf{1.000}        & \textbf{0.995}       & \textbf{1.000}        & \textbf{0.976}       & \textbf{1.000} &  & \textbf{1.000}        & \textbf{0.995}       & \textbf{1.000}        & \textbf{0.976}       & \textbf{1.000} \\ \hline
\multicolumn{12}{c}{C4 (High-entropy Scenario)}                                                                                                   \\ \hline
        & \multicolumn{5}{c}{Sampling}                                    &  & \multicolumn{5}{c}{Beam-search}                                 \\ \cline{2-12} 
Methods & \multicolumn{2}{c}{1\%FPR} & \multicolumn{2}{c}{5\%FPR} & Best  &  & \multicolumn{2}{c}{1\%FPR} & \multicolumn{2}{c}{5\%FPR} & Best  \\ \cline{2-12} 
        & TPR          & F1          & TPR          & F1          & F1    &  & TPR          & F1          & TPR          & F1          & F1    \\ \hline
KGW     & 0.985        & 0.990       & 1.000        & \textbf{0.980}       & 0.992 &  & 0.995       & 0.993       & 1.000        & \textbf{0.980}       & 0.997 \\
SWEET   & 1.000        & 0.997       & 1.000        & 0.978       & 1.000 &  & 0.995        & 0.993       & 1.000        & 0.976       & 0.997 \\
 \rowcolor{gray!20} 
EWD    & \textbf{1.000}        & \textbf{0.997}       & \textbf{1.000}        & 0.978      & \textbf{1.000} &  & \textbf{0.995}        & \textbf{0.993}       & \textbf{1.000}        & 0.976       & \textbf{0.997} \\ \hline
\end{tabular}
\end{table*}
In this section, we first introduce the tasks and datasets used for watermarked text generation. Then, we compare the performance of our EWD against several baselines on those datasets. Next, we analyze the role of our proposed entropy-based weight and study the performance of EWD using different weight computing functions. Finally, we evaluate the detection performance without the original prompt, detection speed and robustness of our EWD.

\subsection{Experiment Settings}

\begin{figure*}[t]
    \centering
    \begin{subfigure}[t]{0.56\textwidth}
        \centering
        \includegraphics[width=0.90\textwidth,height=0.23\textheight]{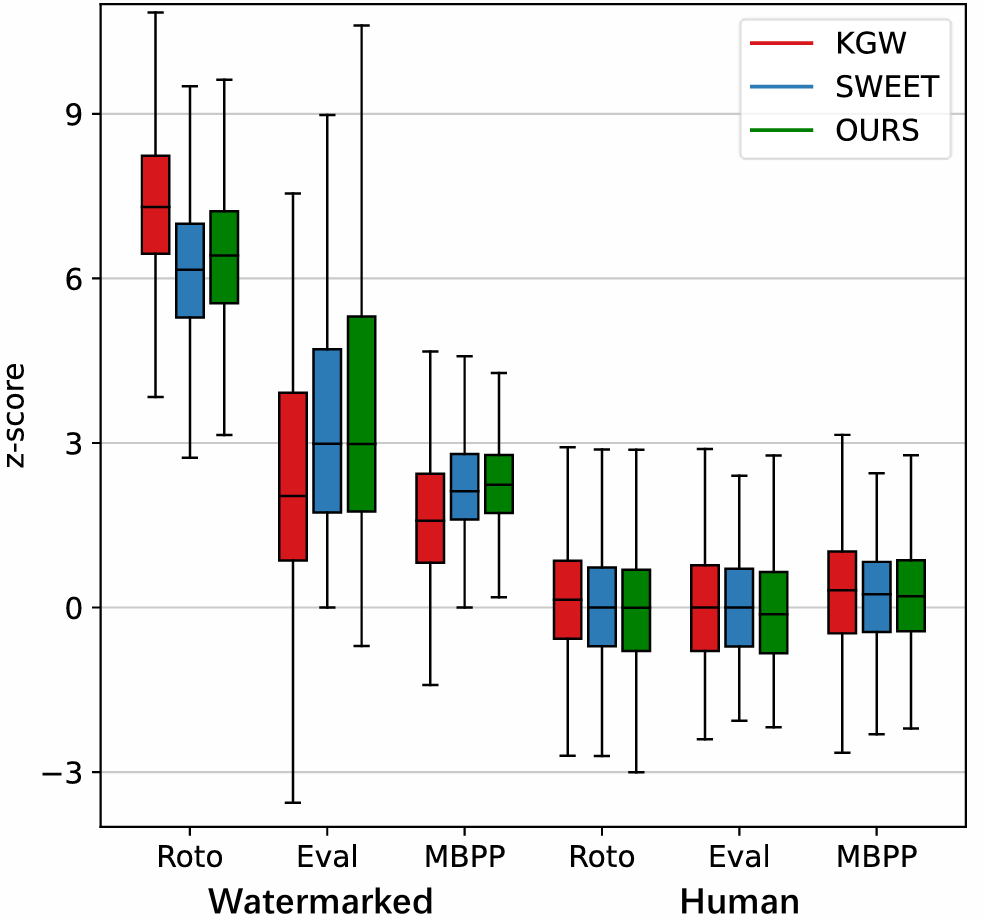}
        \caption{\label{fig:box_a}}
    \end{subfigure}%
    ~\hfill 
    \begin{subfigure}[t]{0.44\textwidth}
        \raggedright
        \includegraphics[width=0.82\textwidth,height=0.23\textheight]{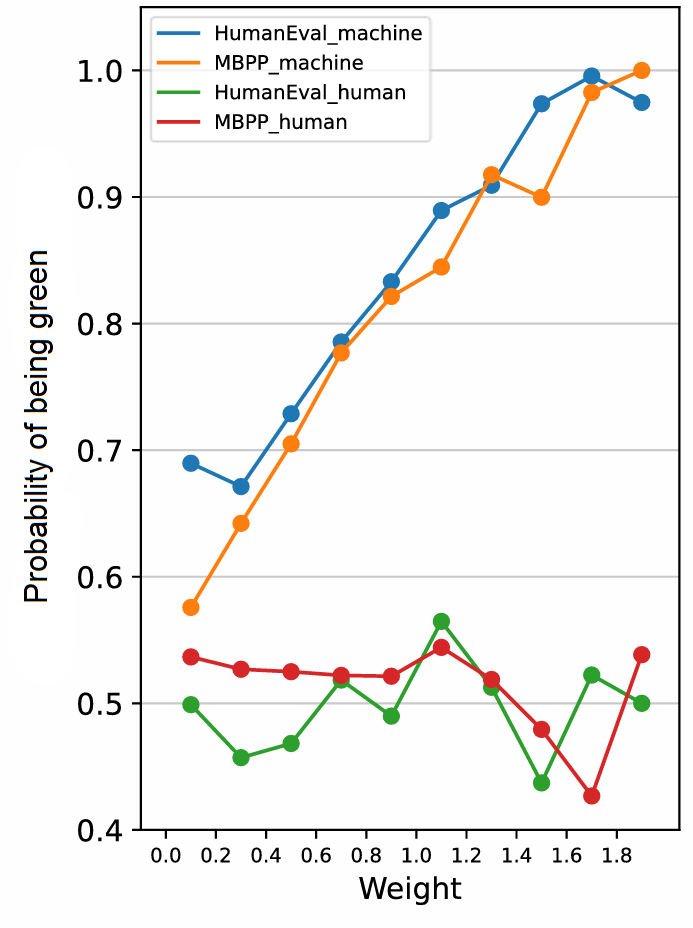}
        \caption{\label{fig:box_b}}
    \end{subfigure}
    \caption{\label{fig:box-fig} Subfigure~(a) shows the z-scores of watermarked and human texts in the Rotowire, HumanEval and MBPP datasets, respectively, each being detected with 3 different methods. Subfigure~(b) shows the relationship between token weights and the probability of being green in both watermarked and human texts.}
\end{figure*}


\begin{figure*}[t]
    \centering
    \begin{subfigure}[t]{0.5\textwidth}
        \centering
        \includegraphics[width=0.94\textwidth,height=0.23\textheight]{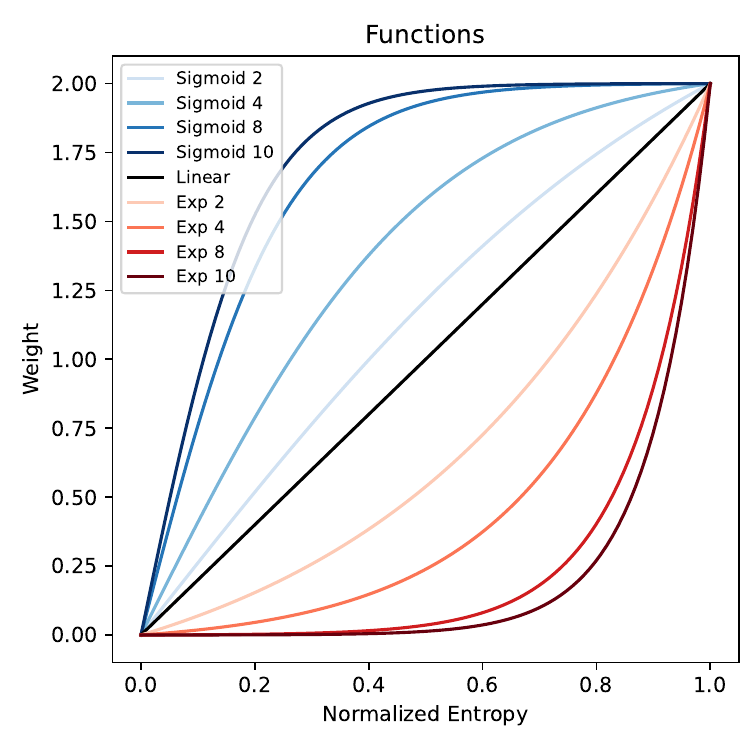}
        \caption{\label{fig:ab_a}}
    \end{subfigure}%
    ~ 
    \begin{subfigure}[t]{0.5\textwidth}
        \raggedright
        \includegraphics[width=0.94\textwidth,height=0.23\textheight]{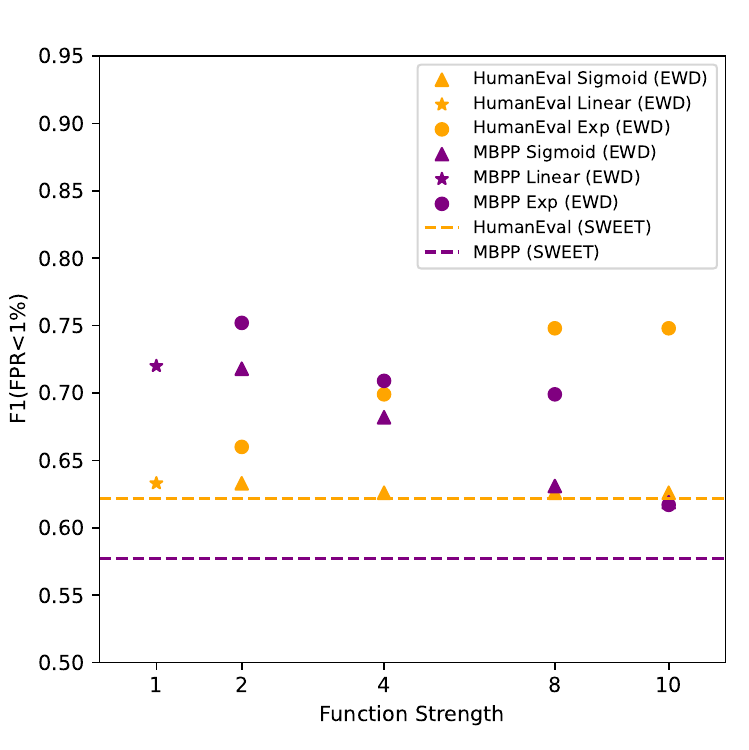}
        \caption{\label{fig:ab_b}}
    \end{subfigure}
    \caption{\label{fig:ablation-fig}We utilize two additional weight functions other than Linear and measured their performance in the code detection datasets. Subfigure~(a) is a visualization of all studied functions, with normalized spike entropy as input. Subfigure~(b) shows each function's detection F1 under 1\% FPR with comparison to the SWEET baseline. Each data point can correspond to the illustrated function on the left figure by looking at its shape and x-axis value.}
\end{figure*}


\textbf{Tasks and Datasets.} There are various text generation tasks in the field of Natural Language Processing, and here we aim to generate watermarked text of both low-entropy and high-entropy distributions. Since our focus is on comparing different detection methods, we utilize the same KGW watermark generation algorithms in all the following tasks and datasets.

For low-entropy scenarios, we target at the task of code generation by adopting two datasets following \citet{lee2023wrote}, HumanEval \cite{chen2021evaluating} and MBPP \cite{austin2021program}. The two datasets contain test cases of python programming problems and corresponding reference answers, which are utilized as human-written samples. Code generation is conducted by StarCoder \cite{li2023starcoder}. During evaluation, the length of both samples are restricted to be at least 15 tokens. 

The first common high-entropy scenario is the news report generation. Our input prompts to the pre-trained OPT-1.3B \cite{zhang2022opt} are from the C4 \cite{raffel2020exploring} dataset. 
Besides, we include another high-entropy data-to-text generation task by utilizing \citet{wu2021text}'s version of the Rotowire dataset \cite{wiseman2017challenges}. The dataset consists of NBA game records, presented as tables, and human-written descriptions. We fine-tune a T5-small \cite{raffel2023exploring} model for text generation and input the table information to the fine-tuned model. We generate 200 samples of length 200 ± 5 tokens for each high-entropy tasks, using binomial sampling and beam search.



\noindent\textbf{Baselines and Evaluation Metrics.} Two detection methods are selected as baselines. The first is KGW detector. The second is SWEET's detector, which excludes low-entropy tokens during detection. Note that SWEET proposes both selective watermark generation and detection, while here we only adopt the detection part. The selection of the entropy threshold follows the instructions in SWEET. For evaluation,  we set the false positive rate (FPR) under 5\%, following \citet{zhao2023provable}'s instruction. We report the true positive rate (TPR) and F1 score for each method.

\noindent\textbf{Hyper-parameters.}
For the KGW generator, $\gamma$ and $\delta$ is set to 0.5 and 2, respectively. Binomial sampling sets the sampling temperature to a common value of 0.7. Beam-search sampling sets the number of beams to 8. To avoid excessive repetition of text, {\fontsize{10pt}{12pt}\selectfont 
\textsf{"no\_repeated\_ngrams"}} is set to 16.

\subsection{Main Results}
Before detection, we measure the generation quality of all generated texts. Specifically, we are able to achieve 31.0\% and 30.4\% in pass@1\cite{chen2021evaluating} for the HumanEval and MBPP dataset, respectively, comparable to \citet{lee2023wrote}'s. For the Rotowire and C4 dataset, our generated output can achieve 7.24 and 3.25 in text perplexity (PPL) using OPT-6.7B \cite{zhang2022opt}.

Table~\ref{table_main} shows that for the low-entropy scenario, watermarked code detection, our detector would achieve the best detection accuracy against other baselines. Specifically, for HumanEval and MBPP, respectively, our method could achieve a 2.5\% and 9.4\% improvement than SWEET in TPR while maintaining a low-than 5\% FPR.

For the high-entropy scenarios using the Rotowire and C4 datasets, the detection performance of our method is overall very similar to other baselines, sometimes even better than some baselines.

\subsection{Empirical Analysis}

\textbf{Impact of Entropy-based Weights.}
\noindent In Figure~\ref{fig:box-fig}, we provide a few visualizations to show the reason why EWD surpasses other baselines in detection performance. Figure~\ref{fig:box_a} shows the z-scores of both watermarked and human texts in different datasets, detected by different methods. Our EWD method would result in overall higher z-scores for watermarked texts, and slightly lower z-scores for human texts. The z-score increase is more significant in low-entropy code generation tasks than in the Rotowire dataset. By enlarging the gap between the z-scores of watermarked and human texts, our detector could better distinguish the watermarked texts. In Figure~\ref{fig:box_b}, we plot the relationship of a token's entropy-based weight and probability of that token being sampled from the green list. In human-written texts, there does not exist a relationship between the probability and token weights, explaining why the z-scores of human text remain low. Meanwhile, in watermarked texts, the probability is proportional to the token weight, which means the tokens assigned with a large weight are very likely to be sampled from the green list and contribute to a larger z-score. 
This proves that our entropy-based weights can help more accurately reflect the watermark level in low-entropy texts.

\textbf{Performance with Different Weight Functions.}
\noindent In the experiments, we utilize a linear function to gain the entropy-based weights for each token. We also study two more monotonic increasing functions, the Sigmoid and the Exponential function. As shown in Figure~\ref{fig:ab_a}, we further change the function strength to 4 different values in each function, which represents how far the function deviates from being linear, for example, "Sigmoid 10" deviates from linear more than "Sigmoid 8". The abovementioned functions are applied in Eq.~(\ref{assignment}) as $f$. Specifically, each function's gradient is different. For our custom Sigmoid functions, the gradient starts off high and gradually decreases, meaning that tokens with relatively low entropy will have significant differences in weights, while tokens with relatively high entropy will exhibit smaller differences. Similarly, for our custom Exponential functions, the gradient starts off low and gradually increases, meaning that tokens with relatively low entropy will have smaller differences in weight, while tokens with relatively high entropy will exhibit larger differences.

Figure~\ref{fig:ab_b} demonstrates that the two listed functions with different strength could surpass the SWEET baseline in terms of detection F1. This reflects the versatility of our method, as it is applicable to more functions.

\begin{table}[]
\centering
\small
\caption{\label{prompt-table}Detection performance of watermarked code using the general prompt instead of original prompts.}
\begin{tabular}{ccc}
\hline
\multicolumn{3}{c}{Detection w/ General Prompt} \\ \hline
Methods      & F1 under 5\%FPR     & Best F1     \\ \hline
KGW         &  0.498                   &  0.741           \\
SWEET       &  0.580                   &  0.762           \\
\rowcolor{gray!20}
EWD         &  \textbf{0.600}                   &  \textbf{0.764}           \\ \hline
\end{tabular}
\end{table}

\textbf{Performance without Original Prompts.} The entropy calculation during detection of our EWD and SWEET is conducted with the original prompts. There exists many real-world scenarios where the original prompts are available, $e.g.$, some programming assignment/exams, in such cases, our detection method would produce the most accurate entropy and detection performance. However, it is possible that the original prompts are not available during detection. Therefore, we design experiments on MBPP dataset where we use a general prompt "{\fontsize{10pt}{12pt}\selectfont 
\texttt{def solution(*args):$\textbackslash$n'''Generate a solution$\textbackslash$n'''}}" to replace the original prompt during entropy calculation. The detection result can be found in Table~\ref{prompt-table}. Even without the original prompt, our EWD could achieve better detection performance against other baselines. However, the detection performance by both SWEET and EWD has dropped compared to using original prompts.

\textbf{Detection Speed.}
\begin{table}[t]
\centering
\small
\caption{\label{complex-table}This table shows the average time taken to generate 200-token texts using KGW generator, as well as the average time taken for detection using different methods, measured in seconds. The experiment is conducted on a NVIDIA RTX A6000 GPU.}
\begin{tabular}{ccc}
\hline
Method & Generation & Detection \\ \hline
KGW    &            & \textbf{0.0391}          \\
SWEET  &  9.489          & 0.0830          \\
\rowcolor{gray!20}
EWD   &            & 0.0831          \\ \hline
\end{tabular}
\end{table}
\noindent
Compared to the KGW detection, both SWEET and our proposed EWD would demand extra time to compute token entropy. Specifically, our detector would also compute weights given the token entropy before calculating the final z-score. According to Table~\ref{complex-table}, the detection time per text is 0.0391, 0.0830 and 0.0831 seconds for KGW, SWEET and our EWD, respectively. In spite of being double the value of KGW, our detector remains highly efficient, and the difference with SWEET is virtually insignificant.

\textbf{Performance against the Back-translation Attack.}
\begin{table}[t]
\small
\centering
\caption{\label{robust-table}Detection performance of back-translated watermarked texts using different detection methods.}
\begin{tabular}{ccccc}
\hline
\multicolumn{5}{c}{Detection w/ Back-translation}                                         \\ \hline
\multirow{2}{*}{Methods} & \multicolumn{2}{c}{1\%FPR} & \multicolumn{2}{c}{5\%FPR} \\ \cline{2-5} 
                         & TPR          & F1          & TPR          & F1          \\ \hline
KGW        &    0.894   &   0.940   & \textbf{0.950} & \textbf{0.952}               \\
SWEET          &   0.890   &   0.938   &   0.930   &   0.940           \\
\rowcolor{gray!20}
EWD        & \textbf{0.900} & \textbf{0.943} &   0.942   &   0.947        \\ \hline
\end{tabular}
\end{table}
Watermarked texts usually would be edited before being detected, which would remove part of the watermark and cause detection performance drop. Here we also evaluate the baseline methods and our EWD under this setting by utilizing back-translation as the attack to remove watermark. Specifically, we firstly generate watermarked texts using the C4 dataset, then translate them from English to French, and later back to English again for detection. In Table~\ref{robust-table}, our EWD would achieve the best detection accuracy under 1\% FPR and similar results with KGW under 5\% FPR. 

\section{Conclusion}
In this work, we propose a text watermark detection method called EWD that fully considers the influence of token entropy. We first establish the positive influence-entropy relation in the detection process, then utilize entropy for weight customization to fully reflect the proposed relation. We provide theoretical analysis on the detection properties of different methods in low-entropy scenarios. Experiments validate the detection performance of our method in the low-entropy code generation task with a number of weight functions. Furthermore, in high-entropy scenarios, our method would achieve similar performance as previous works in detection accuracy, detection speed and robustness.

\section*{Limitations}
Our method mainly includes two limitations. The first one is that the low-entropy datasets tested are limited. We utilize two code generation datasets for evaluation. In the future, we aim to include more low-entropy tasks and datasets. 

Secondly, our method should be theoretically effective for various types of watermark methods, such as token sampling-based methods. In the future, we aim to implement our method on more types of text watermarking frameworks.

\section*{Acknowledgement}
We sincerely appreciate the insightful comments from all reviewers. Their help has greatly contributed to the improvement of this work.

The work described in this paper was partially supported by the Research Grants Council of the Hong Kong Special Administrative Region, China (CUHK 14222922, RGC GRF 2151185).
\newpage

\bibliography{acl_latex}
\clearpage
\appendix
\section{Proof of Type-II Error for EWD}
\label{appendix:1}
Here we demonstrate how we obtain the theoretical Type-II error for EWD. First of all, we attempt to calculate the theoretical $\mu$ and $\sigma^2$ based on Eq.~(\ref{kgw_prob}). The theoretical $\mu$ is the weighted sum of green tokens in the watermarked text, using ${|s|_G}_k$ to represent the weighted number of green tokens from each token $k$,thus we can obtain:
\begin{alignat}{2}
\mu &= \sum_{k}{|s|_G}_k,  \\
    &= \sum_{k}^{}\mathbb{P}[k\in G]\cdot W(k),  \\
    &= T\mathbb{E}\frac{\gamma \alpha SE}{1+(\alpha-1)\gamma}(SE-C_0),  \\
    &= T\frac{\gamma \alpha}{1+(\alpha-1)\gamma}(\mathbb{E}[SE^2]-C_0\mathbb{E}[SE]).
\end{alignat}

Given the simulated power law of spike entropy, we can obtain $E[SE^2]$ and $E[SE]$ as 0.377 and 0.608, respectively, producing $\mu$ equal to 5.79. And for $\sigma^2$, we first obtain the variance for each token $k$:
\begin{alignat}{2}
    {\sigma_k}^2 &= \mathbb{E}[({|s|_G}_k-\mathbb{E}[{|s|_G}_k])^2],    \\
    &= \mathbb{E}[{|s|_G}_k^2]-\mathbb{E}[{|s|_G}_k]^2, \\
    &= \mathbb{P}[k\in G]\cdot W(k)^2-\mathbb{P}[k\in G]^2 \cdot W(k)^2, \\
    &= W(k)^2 \cdot (\mathbb{P}[k\in G])(1-\mathbb{P}[k\in G]).
\end{alignat}

Therefore, if we represent $\frac{\gamma \alpha}{1+(\alpha-1)\gamma}$ by $C_1$, the variance of the whole text can be formulated as:
\begin{alignat}{2}
    \sigma^2 &= \sum_{k}W(k)^2 \cdot (\mathbb{P}[k\in G])(1-\mathbb{P}[k\in G]),    \\
    &= T\mathbb{E}[W(k)^2C_1SE(1-C_1SE)], \\ 
    &= TC_1\mathbb{E}[(SE-C_0)^2SE(1-C_1SE)], \\
    &= TC_1(-C_1\mathbb{E}[SE^4]+(2C_0C_1+1)\mathbb{E}[SE^3] \nonumber \\
    &\ -(2C_0+{C_0}^2C_1)\mathbb{E}[SE^2]+{C_0}^2\mathbb{E}[SE]).
\end{alignat}

Given the simulated power law of spike entropy, we can obtain $E[SE^4]$ and $E[SE^3]$ as 0.24 and 0.158, respectively, producing $\sigma$ equal to 0.56. When the detection threshold $z'$ is set as 2 (corresponding to $|\tilde{s}|_G$ = 5.55), the probability of $|s|_G \sim \mathcal{N}(5.79, 0.31)$ falling below 5.55 is 33.4\%, $i.e.$, the Type-II error.

\end{document}